\documentclass[10pt,twocolumn,letterpaper]{article}

\usepackage[pagenumbers]{iccv} 

\definecolor{iccvblue}{rgb}{0.21,0.49,0.74}
\usepackage[pagebackref,breaklinks,colorlinks,allcolors=iccvblue]{hyperref}

\usepackage{multirow}
\definecolor{mycolor_blue}{HTML}{E7EFFA}
\definecolor{mycolor_green}{HTML}{E6F8E0}
\definecolor{mycolor_gray}{HTML}{ECECEC}
\definecolor{mycolor_red}{HTML}{FFE6E6}
\definecolor{mycolor_yellow}{HTML}{FFFFCC}
\definecolor{mycolor_purple}{HTML}{E6E6FF}
\usepackage{xcolor} 
\usepackage{graphicx}
\usepackage{colortbl}
\usepackage{tcolorbox}  
\usepackage{titletoc}

\def\paperID{****} 
\def\confName{ICCV}
\def\confYear{2025}

\title{Personalized Text-to-Image Generation with Auto-Regressive Models}

\author{
  \bf{Kaiyue Sun}\textsuperscript{1} \quad 
  \bf{Xian Liu}\textsuperscript{2}  \quad 
  \bf{Yao Teng}\textsuperscript{1} \quad 
  \bf{Xihui Liu}\textsuperscript{1} \\
\textsuperscript{1}The University of Hong Kong  \quad
\textsuperscript{2}The Chinese University of Hong Kong  \quad 
\\
\\
Code: \href{https://github.com/KaiyueSun98/T2I-Personalization-with-AR}{https://github.com/KaiyueSun98/T2I-Personalization-with-AR}
}
\begin{document}

\twocolumn[{%
\renewcommand\twocolumn[1][]{#1}%
\maketitle
\vspace{-2em}
\includegraphics[width=\linewidth]{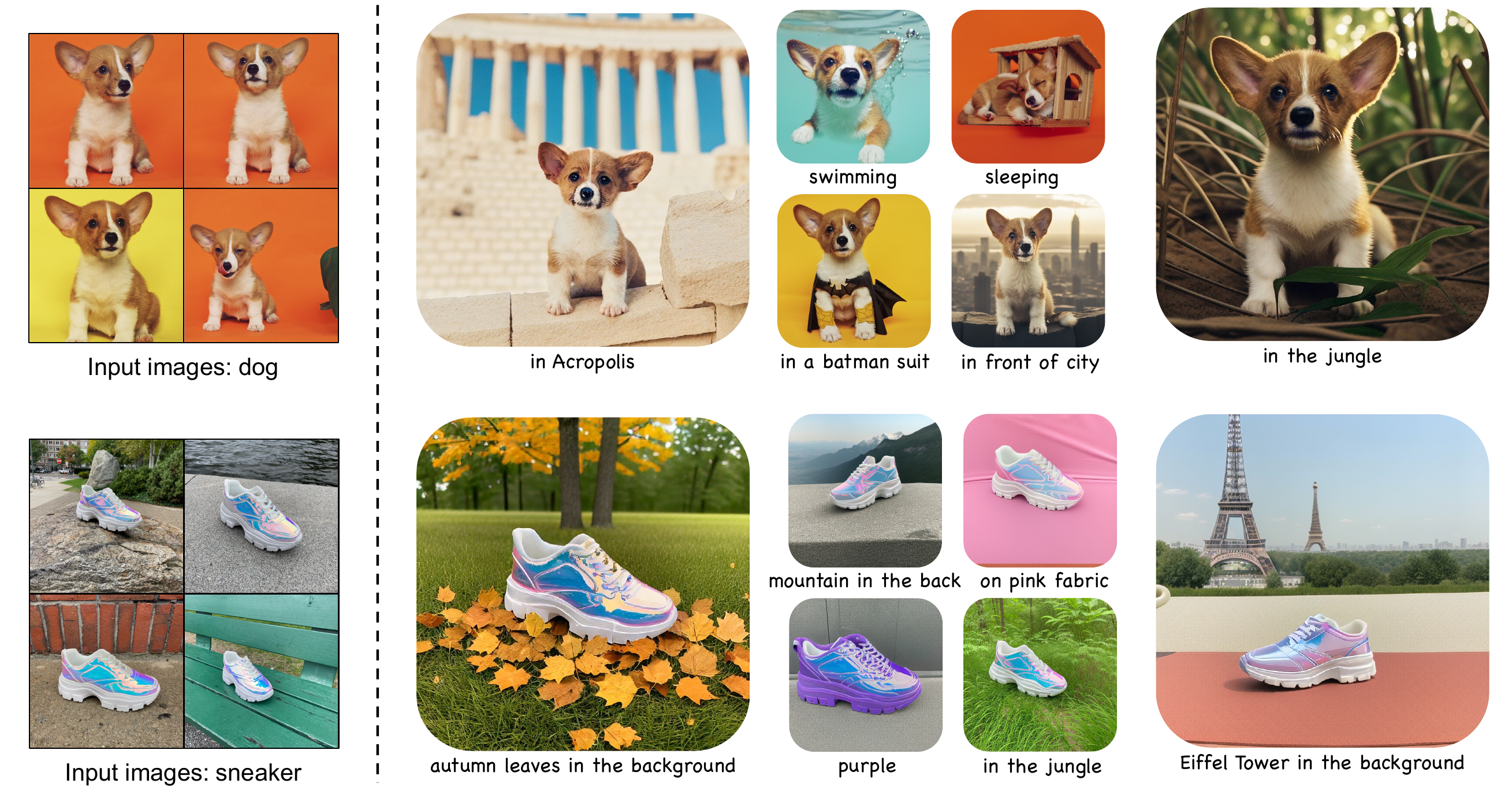}
\captionof{figure}{\textbf{Overview.} Using only a few reference images (typically 3-5) of a subject (left), we fine-tune an auto-regressive model to generate personalized images of the subject in diverse contexts (right), guided by text prompts. \vspace{2em}
}
\label{fig:teaser}
}]

\maketitle
\begin{abstract}
Personalized image synthesis has emerged as a pivotal application in text-to-image generation, enabling the creation of images featuring specific subjects in diverse contexts. While diffusion models have dominated this domain, auto-regressive models, with their unified architecture for text and image modeling, remain underexplored for personalized image generation. This paper investigates the potential of optimizing auto-regressive models for personalized image synthesis, leveraging their inherent multimodal capabilities to perform this task. We propose a two-stage training strategy that combines optimization of text embeddings and fine-tuning of transformer layers. Our experiments on the auto-regressive model demonstrate that this method achieves comparable subject fidelity and prompt following to the leading diffusion-based personalization methods. The results highlight the effectiveness of auto-regressive models in personalized image generation, offering a new direction for future research in this area.
\end{abstract}
    
\section{Introduction}
\label{sec:intro}
The rapid advancement of text-to-image generation models has revolutionized the field of computer vision, enabling the creation of highly realistic and diverse images from textual descriptions. Among the various applications of these models, personalized image synthesis—generating images of specific subjects in new contexts—has garnered significant attention. This capability is particularly valuable for applications in digital art, advertising, and virtual reality, where the ability to seamlessly integrate personalized content into diverse scenes is crucial.

While diffusion models have been at the forefront of personalized image generation, auto-regressive models, which employ a unified architecture for text and image modeling, have not been extensively explored for this task. Auto-regressive models~\cite{Dalle-2,yu2022parti,liu2024lumina-mgpt,sun2024autoregressive,wang2024emu3} have demonstrated remarkable success in text-to-image generation by predicting image tokens sequentially. However, their potential for personalized image synthesis remains largely untapped. This paper aims to investigate the adaptation of auto-regressive models for personalized image generation.

We propose a novel two-stage training strategy that firstly optimizes text embeddings and then fine-tunes transformer layers together. 
Our experiments on the Lumina-mGPT 7B model~\cite{liu2024lumina-mgpt} show that this approach outperforms existing optimization-based techniques like Textual inversion~\cite{gal2022textual} and shows comparable performance with Dreambooth~\cite{ruiz2023dreambooth} in terms of subject fidelity and prompt following. The results underscore the potential of auto-regressive models in personalized image generation and pave the way for future research in this domain.

This work explores the potential of auto-regressive models for personalized image synthesis, adapting them to meet the specific demands of text-to-image generation. Our findings suggest that auto-regressive models, when properly optimized, can achieve competitive performance in personalized image generation, offering a promising alternative to diffusion-based approaches.

\section{Related Work}

\subsection{Personalized Image Synthesis.}
Personalized image synthesis focuses on generating images of specific subjects in novel contexts while seamlessly integrating them into the new scene. 
The advent of diffusion models~\cite{ho2020denoising, song2019generative, song2021scorebased} has significantly advanced text-guided image generation, offering remarkable flexibility to personalize text-to-image diffusion models~\cite{ruiz2023dreambooth,gal2022textual,avrahami2023break,han2023svdiff,kumari2023multi,tewel2023key,gu2023mix,kim2024selectively,
wei2023elite,li2023blip,ma2024subject,ye2023ip-adapter,gal2023encoder,jia2023taming,shi2024instantbooth,
patel2024lambda,mao2024realcustom++,chen2023subject,pan2023kosmos}. These models can customize single or multiple subjects using a few reference images or even a single image.  
Current diffusion-based personalization models can be broadly divided into two categories: optimization-based models and tuning-free models.
Optimization-based models~\cite{ruiz2023dreambooth,gal2022textual,avrahami2023break,han2023svdiff,kumari2023multi,tewel2023key,gu2023mix,kim2024selectively}, such as Textual Inversion~\cite{gal2022textual} and DreamBooth~\cite{ruiz2023dreambooth}, optimize embeddings of text tokens or fine-tune the backbone to implant a new subject into the model’s output domain. In contrast, tuning-free models~\cite{wei2023elite,li2023blip,ma2024subject,ye2023ip-adapter,gal2023encoder,jia2023taming,shi2024instantbooth,
patel2024lambda,mao2024realcustom++,chen2023subject} pre-train a separate encoder to extract the subject features, enabling zero-shot personalized image generation without per-subject optimization. 

Recent advancements in unified multimodal models~\cite{xiao2024omnigen,sun2024emu2,ge2024seedx} have expanded personalized image synthesis into broader multimodal tasks. These unified generative models typically rely on carefully curated datasets for instruction tuning or large-scale training. However, when personalized image generation is included as one of many tasks for these models, its performance often lags behind models specially designed for this task.

\subsection{Auto-regressive Image Generation.}
Auto-regressive models generate data through next token prediction. Early auto-regressive image generation models like PixelCNN~\cite{van2016pixelcnn} and PixelRNN~\cite{van2016pixelrnn} operated at the pixel level, but their computational demands became prohibitive due to the high dimensionality of images. The introduction of VQVAE~\cite{van2017vqvae} addressed this by compressing images into discrete tokens and enabled a two-stage approach: image tokenization followed by auto-regressive modeling of token distributions. This framework revitalized auto-regressive image generation.

Building on this, models like DALL-E~\cite{Dalle-2} and CogView~\cite{ding2021cogview} extended auto-regressive methods to text-to-image generation. They compress images into tokens with an autoencoder. By concatenating the text tokens and image tokens, they train a decoder-only transformer to predict the next image token.
Parti~\cite{yu2022parti} and LlamaGen~\cite{sun2024autoregressive} employ separate text encoders to inject the text features into the auto-regressive decoder.

Chameleon~\cite{team2024chameleon} advances the framework by enabling mixed-modal generation of interleaved text and image sequences using a unified transformer architecture, eliminating the need for separate text encoders. Lumina-mGPT~\cite{liu2024lumina-mgpt} inherits Chameleon's tokenizers and trains an auto-regressive transformer from scratch, enabling image generation and other vision-centric tasks. These developments highlight the versatility and scalability of auto-regressive models in multimodal generation.

Diffusion models and auto-regressive models differ fundamentally in their paradigms. Diffusion models excel in iterative refinement, while auto-regressive models employ a unified architecture for text and image modeling, making them inherently compatible with multimodal tasks. Despite this, auto-regressive models specifically tailored for personalized image generation remain unexplored. This work aims to investigate the potential of optimizing auto-regressive models for personalized image generation.

\section{Preliminaries}

\subsection{Personalizing Text-to-Image Models via Optimization}
\noindent \textbf{Textual Inversion.} Textual Inversion~\cite{gal2022textual} proposes a personalization method by creating a new ``pseudo-word'' (e.g., $S_*$) within the text embedding space of a text-to-image diffusion model. Using just 3-5 images of a specific subject provided by the user, this method optimizes the embedding vector corresponding to the pseudo-word to represent that subject.
This word can then be used to compose natural language prompts, such as ``a $S_*$ on the beach'', to generate personalized images in novel contexts.

\noindent \textbf{DreamBooth.} Instead of a ``pseudo-word'', DreamBooth~\cite{ruiz2023dreambooth} opts to optimize a unique identifier ``[V]'' that precedes the subject class (for example, ``a [V] cat/dog/toy on the beach''). This approach helps to link the prior knowledge of the class with the subject, thereby reducing training time. However, using the class name can lead to a gradual loss of the model's broader semantic knowledge during fine-tuning, a phenomenon known as language drift. To address this issue, a class-specific prior preservation loss is introduced to retain the model's ability to generate diverse instances of the class.

These optimization-based approaches are implemented and
proved to be effective on text-to-image diffusion models.
They can effectively perform various personalization tasks, including subject recontextualization, text-guided view synthesis, and artistic rendering. 

In this paper, we explore the adaptation of these optimization-based personalization techniques to auto-regressive models and offer insights into the finetuning of auto-regressive models.

\subsection{Text-to-Image Generation via Next-Token Prediction}
Auto-regressive text-to-image models generate images in three steps. First, a tokenizer converts the input text into a sequence of discrete tokens, which are transformed into vector embeddings. These text embeddings, denoted as $c$, are then fed into an auto-regressive transformer that outputs logits $l_t$. 
The logits are converted into probabilities where the next image token $x_t$ is sampled. The newly sampled token is concatenated with the preceding tokens to predict the subsequent token.
Finally, an image decoder translates the complete sequence of tokens $x = (x_1, x_2, \ldots, x_T)$ into image pixels.

\noindent \textbf{Training objective.}
During training, the auto-regressive transformer models the conditional probability 
$p\left(x_t \mid x_1, x_2, \ldots, x_{t-1}, c \right)$
of the sequential tokens using the standard next-token prediction objective.
We denote $x_{1\sim t-1} = \{ x_1, x_2, \ldots, x_{t-1}\}$ , the model predicts the next token $x_t \in V$, where $V$ denotes the vocabulary. The loss function $f$ for a single prediction can be written as follows:
\begin{equation}
    L(\theta) = f\left( y_t , p_{\theta}\left(x_t \mid x_{1\sim t-1}, c \right)\right),
\label{equ1}
\end{equation}
\begin{equation}
p_{\theta}\left(x_t \mid x_{1\sim t-1}, c \right) = \text{Softmax}(l_t),
\label{equ2}
\end{equation}

\noindent where
$L(\theta)$ is the loss, parameterized by the model parameters $\theta$ and loss function $f$. In image generation, we predict the tokens from the image split of the total vocabulary.
$y_t$ represents the target label of the next token, which is derived by tokenizing the ground-truth image associated with the input text. 
$f$ is cross-entropy loss.

\noindent \textbf{Classifier-free guidance}. Most auto-regressive text-to-image models adopt classifier-free guidance (CFG) to enhance the quality of generated images. When generating an image token, the logits processed by CFG, denoted as $l_{t\_cfg}$ are formulated as follows: 
\begin{equation}
l_{t\_cfg} = s(l_{t}-l_{t}^{'}) + l_{t}^{'},
\label{equ3}
\end{equation}
where $l_{t}$ represents the original logits that are conditioned on the complete input text; $l_{t}^{'}$ refers to the context-independent logits, which are independent of any prior token. The variable $s$ denotes the guidance scale used in classifier-free guidance.

In this work, we aim to personalize the auto-regressive model by finding a unique identifier of a subject and fine-tuning the transformer layers using a set of reference images of that subject.

\section{Method}
\begin{figure}[ht]   
  \centering
   \includegraphics[width=1.0\linewidth]{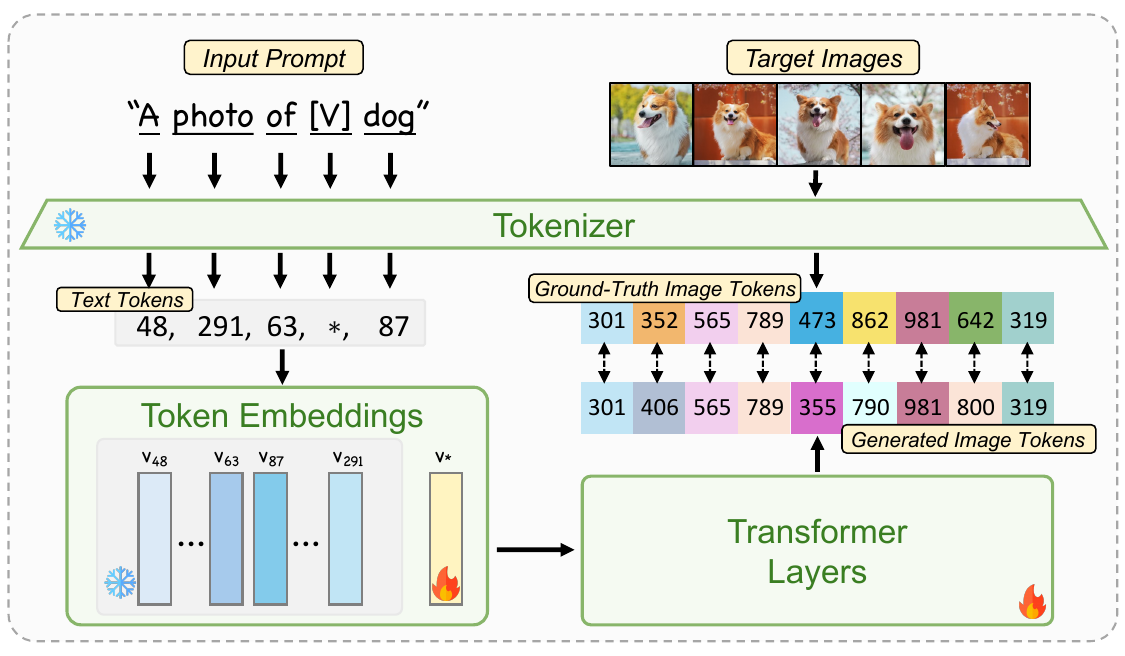}
   \caption{\textbf{Overview of Fine-tuning.}
We fine-tune a text-to-image auto-regressive model using 3-5 input images, each paired with a text prompt that includes a unique identifier and the subject class name (e.g., ``A photo of [V] dog''). The process involves two stages: first, we fine-tune the text embedding for the identifier [V], and second, we additionally fine-tune the transformer layers to enhance the model's performance.}
   \label{fig:architecture}
   \vspace{-10pt}
\end{figure}

Personalizing a text-to-image diffusion model generally involves two strategies. The first strategy is to associate a unique text embedding with the subject. This text embedding can either represent the subject as a whole or serve as an adjective describing the subject class. However, because the number of parameters for a text embedding is limited, personalized images often struggle to capture all the essential features of the subject. To effectively embed the subject's appearance in the model, fine-tuning of the model parameters is usually required. Figure~\ref{fig:architecture} shows the overview of our fine-tuning strategy.

In this section, we present our method for personalizing an auto-regressive model and explain the rationale behind our choices.

\subsection{Optimizing Text Embeddings}
We generally follow the DreamBooth~\cite{ruiz2023dreambooth} approach to optimize a text embedding for a specific subject. We introduce a placeholder word [V], to represent the unique identifier of the new subject we wish to learn. The input text that includes the identifier [V] and the subject class name is then converted to tokens. 
We replace the embedding associated with the token for [V] with a new randomly initialized embedding, denoted as $v_*$.
With a small set of reference images (e.g. 3-5) of the subject in various backgrounds or poses, we optimize $v_*$ based on the cross-entropy loss defined in Equation \ref{equ1}. For the input text, we use the templates provided by Textual Inversion~\cite{gal2022textual}, which contain neural context such as ``A photo of [V] [class\_name]'', ``A rendition of [V] [class\_name]''. 
Our optimization goal can thus be defined as follows:
\begin{equation}
    v_* = \mathop{\arg\min}_v f\left( y_t , p_{\theta}\left(x_t \mid x_{1\sim t-1}, c \right)\right), \forall t
\label{equ4}
\end{equation}

\noindent It is expected to encourage embedding $v_*$ to learn the common features in the reference images while discard elements that are unique to each image, such as the background.

\noindent \textbf{Per-image tokens}
Textual Inversion~\cite{gal2022textual} introduces a scheme that inserts a unique token for each reference image in the text template.
Given $n$ reference images, the method associates each image with a unique placeholder $S_i$, where $i=1,\ldots n$. The embeddings corresponding to these placeholders are optimized to capture the distinct features of each individual image, while minimizing the influence of non-common elements on the universal placeholder [V]. 
  
\noindent Textual Inversion~\cite{gal2022textual} composes input prompts using the format ``A photo of [V] [class\_name] with $S_i$''.
We have observed that auto-regressive models are sensitive to specific wording. 
When prompts use different prepositions or structures during inference, the quality of the generated images tends to decrease.
To avoid this issue, we use the format ``A photo of [V] [class\_name] $S_i$'', where $S_i$ can represent any form of phrase.

\noindent Additionally, we find that introducing the per-image tokens in all training prompts affects the quality of the generated images when we attempt to reconstruct the subject using the prompt ``A photo of [V] [class\_name]''. However, removing all per-image tokens makes it more challenging to eliminate content that is irrelevant to the subject. To strike a balance, we opt for a compromise strategy that incorporates both formats. This strategy improves the robustness of the subject learning process. In our implementation, we use a 1:1 ratio for the two formats. 

\subsection{Fine-tuning Transformer Layers}

We conduct experiments using the Lumina-mGPT 7B model~\cite{liu2024lumina-mgpt}. We have observed that the generated images fail to accurately replicate the reference subject if we optimize the text embeddings only. Additionally, when optimizing the text embeddings on a single data point, the model does not overfit; instead, after a slight decrease in the loss, it stabilizes around a specific level. Given the limited capacity of text embeddings, fine-tuning the auto-regressive transformer becomes necessary to effectively implant the subject into the model's output domain.

\noindent \textbf{Two-stage training.}
DreamBooth~\cite{ruiz2023dreambooth} fine-tunes the layers conditioned on the text embeddings and the diffusion UNet simultaneously. In our experiments, we find that when fine-tuning the text embeddings and transformer layers together, the text embeddings cannot get fully trained. If we revert to the original transformer layers during inference, the text embeddings alone fail to convey any meaningful content. 

\noindent To address this issue, we devise a two-stage training strategy. In the first stage, we fully optimize the text embeddings, and in the second stage, we fine-tune the transformer layers to maximize the subject fidelity. This two-stage approach is mutually beneficial: the first stage stabilizes the training and reduces the effort needed in the second stage, while the second stage compensates for any defects from the first stage due to its inherent limitations.

\subsection{Implementation details} 
Unless otherwise specified, we maintain the original hyper-parameter settings of Lumina-mGPT~\cite{liu2024lumina-mgpt}. The text embeddings are initialized randomly. Our experiments use a batch size of 1, which corresponds to approximately 8 text-image pairs. In the first stage, we set the learning rate for optimizing the text embeddings to 0.01, training for $\sim$1500 steps. 
In the second stage, $\sim$
70 steps with a learning rate of $5 \times 10^{-6}$ is enough for full fine-tuning.

\noindent We find that these training parameters work well for most cases. For some common subjects (e.g., dog, cat), satisfactory results can be achieved with fewer training steps in both stages. However, for some challenging subjects, more training steps in the second stage are required to obtain better results. 
The first stage takes about 15 minutes, and the second stage takes only 2 minutes on a single H100 GPU.

\section{Experiments}

\subsection{Dataset and Evaluation}
We evaluate our model's personalization capability on Dreambench~\cite{ruiz2023dreambooth}, which provides a dataset consisting of 30 subjects, each with 4-6 images. These subjects are divided into two groups: 21 objects and 9 live subjects/pets. Each subject is tested on 25 prompts, which include scenarios such as re-contextualization, accessorization, and property modification. Their purpose is to assess whether the key features of the subject can be preserved under different semantic modifications while the generated image adheres to the prompt.
Following Dreambench~\cite{ruiz2023dreambooth}, we employ DINO~\cite{caron2021emerging_dino} and CLIP-I~\cite{radford2021learning} to assess subject fidelity, and CLIP-T~\cite{radford2021learning} to measure the prompt following. For evaluation, we generate images using a fixed CFG value of 4.0 and an image top-k value of 2000.

Subject fidelity is computed as follows: for each generated image of a subject, we calculate the average similarity between that image and all reference images. This process is repeated for all 25 images generated from the 25 prompts, and the results are averaged to obtain the subject fidelity score for that subject. The overall subject fidelity score for the model is then derived by averaging the scores across all 30 subjects in the dataset. For prompt following, the score for each generated image is calculated by computing the cosine similarity between CLIP~\cite{radford2021learning} embeddings of the image and the corresponding prompt.

\subsection{Quantitative Results}
Table~\ref{tab:benchmark} presents the evaluation results of various models on Dreambench~\cite{ruiz2023dreambooth}. By fine-tuning the auto-regressive model of Lumina-mGPT~\cite{liu2024lumina-mgpt} using our method, it outperforms Textual Inversion~\cite{gal2022textual}, Re-Imagen~\cite{chen2022re}, and zero-shot BLIP-Diffusion~\cite{li2023blip} in both subject fidelity (Dino and CLIP-I) and prompt following (CLIP-T). Additionally, it achieves comparable results to stable diffusion-based DreamBooth~\cite{ruiz2023dreambooth} and fine-tuned BLIP-Diffusion~\cite{li2023blip} in DINO. Notably, our method achieves the highest prompt following score among all the models listed. These findings demonstrate that auto-regressive models can be fine-tuned to incorporate new concepts without compromising their original generation capabilities.

\begin{table}[h]
\centering
\resizebox{\columnwidth}{!}{
  \begin{tabular}{lcccc}
    \toprule
    Method & DINO $\uparrow$ & CLIP-I $\uparrow$ & CLIP-T $\uparrow$\\
    \midrule
    Real Images & 0.774 & 0.885 & N/A \\
    \midrule
    Textual Inversion~\cite{gal2022textual} & 0.569 & 0.780 & 0.255 \\
    Re-Imagen~\cite{chen2022re} &  0.600 &  0.740 & 0.270 \\
    DreamBooth (Stable Diffusion)~\cite{ruiz2023dreambooth} & 0.668 & 0.803 & 0.305 \\
    DreamBooth (Imagen)~\cite{ruiz2023dreambooth} & \textbf{0.696} & \textbf{0.812} & 0.306 \\
    BLIP-Diffusion (zero-shot)~\cite{li2023blip} & 0.594 &  0.779 &  0.300\\
    BLIP-Diffusion (fine-tune)~\cite{li2023blip} & 0.670 & 0.805 &  0.302\\
    \midrule
    \textbf{Ours (Lumina-mGPT~\cite{liu2024lumina-mgpt})} & 0.671 & 0.785 & \textbf{0.314} \\
    \bottomrule
  \end{tabular}
}
\caption{\textbf{Quantitative results comparison on Dreambench~\cite{ruiz2023dreambooth}.} We show subject fidelity (DINO, CLIP-I) and prompt following (CLIP-T) scores across different models. For all three metrics, the scores range from 0 to 1, where a higher score indicates better performance. The bold values highlight the highest score achieved.}
\label{tab:benchmark}
\end{table}

\subsection{Qualitative Results}
\begin{figure*}[ht]   
  \centering
   \includegraphics[width=1.0\linewidth]{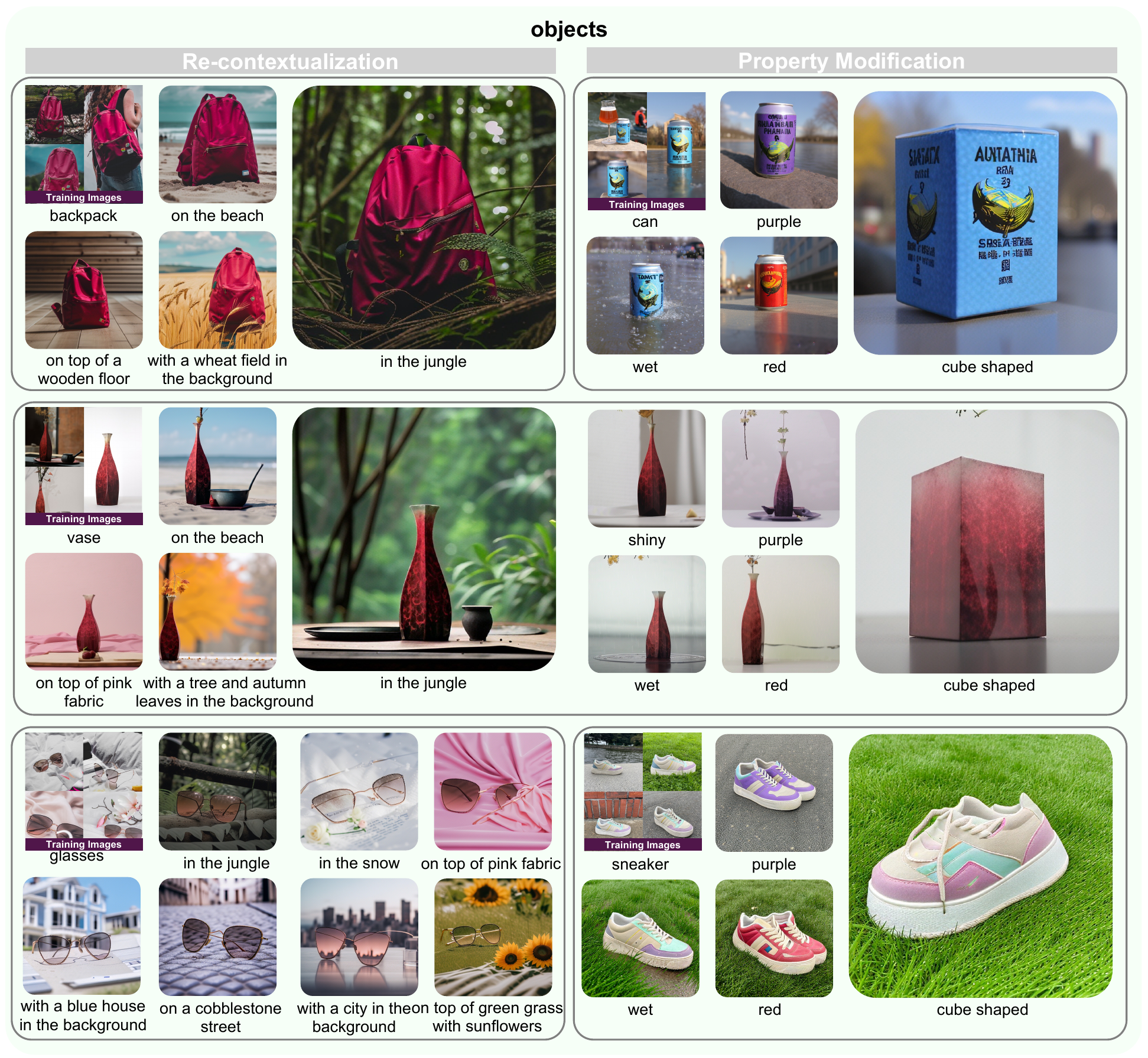}
   \vspace{-10pt}
   \caption{\textbf{Qualitative results.} We generate images of personalized objects to showcase the generative capabilities of re-contextualization and property modification.
   }
   \label{fig:object}
\end{figure*}

\begin{figure*}[ht]   
  \centering
   \includegraphics[width=1.0\linewidth]{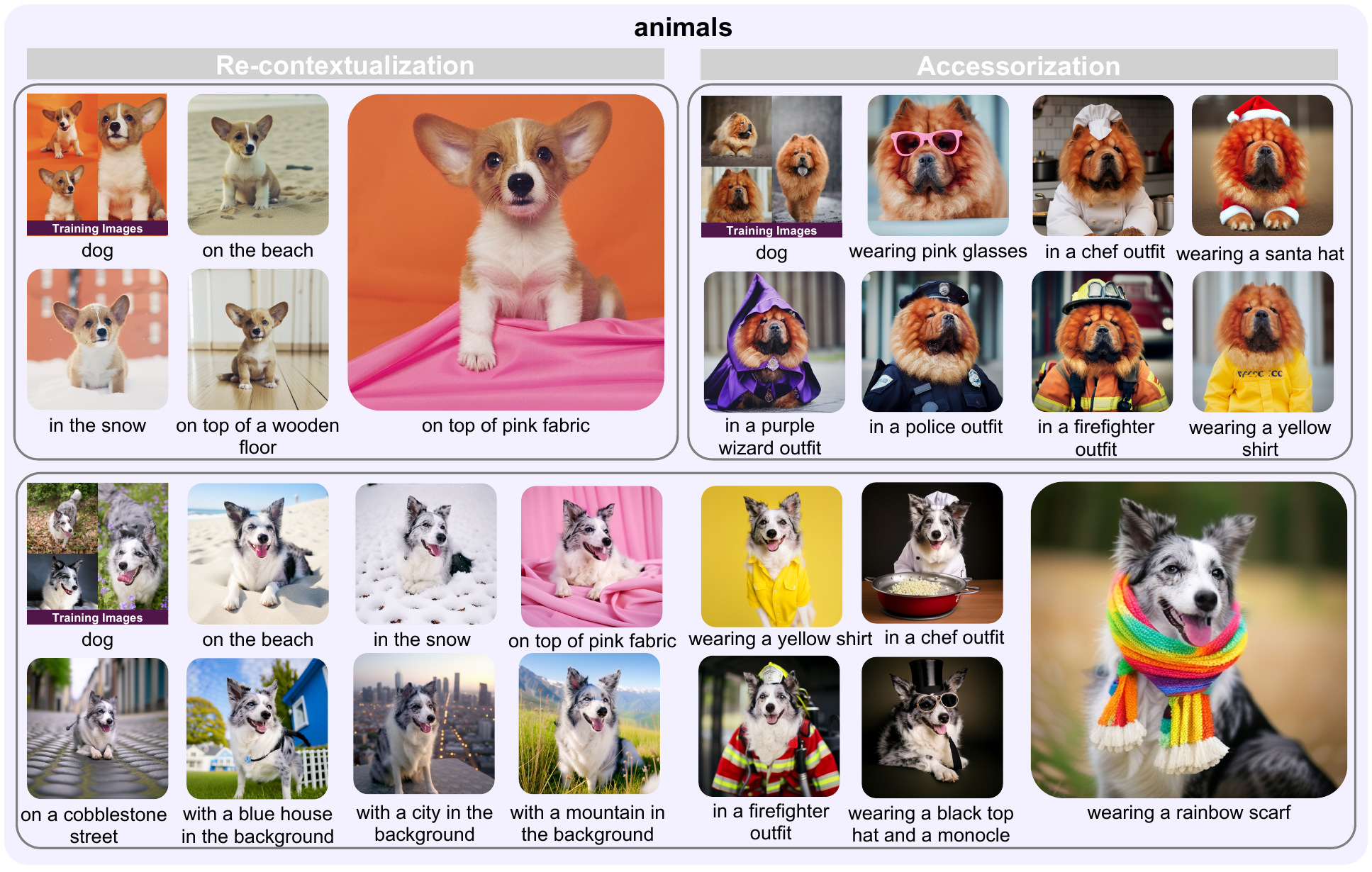}
   \vspace{-10pt}
   \caption{\textbf{Qualitative results.} We generate images of personalized animals to showcase the generative capabilities of re-contextualization and accessorization.
   }
   \label{fig:animal}
\end{figure*}

In Figure~\ref{fig:object} and Figure~\ref{fig:animal}, we present qualitative generation results of our model. The re-contextualization examples demonstrate the model's ability to accurately reproduce the subject's appearance while merging it into the new backgrounds. Furthermore, the model can accurately modify the color and shape properties of the subject, even in challenging cases such as ``cube-shaped''. This indicates that the model not only learns new concepts, but also effectively decomposes and recomposes them with its prior knowledge. 
In accessorization examples, the model can seamlessly integrate subjects with outfits, demonstrating its ability to understand the structure and meaning of the subject rather than merely replicating its appearance.
These results validate the model's strong ability to follow prompts and maintain high subject fidelity, as reflected in the quantitative evaluation.

\subsection{Ablation Studies}

\begin{figure}[htbp]
   \centering
   \includegraphics[width=\linewidth]{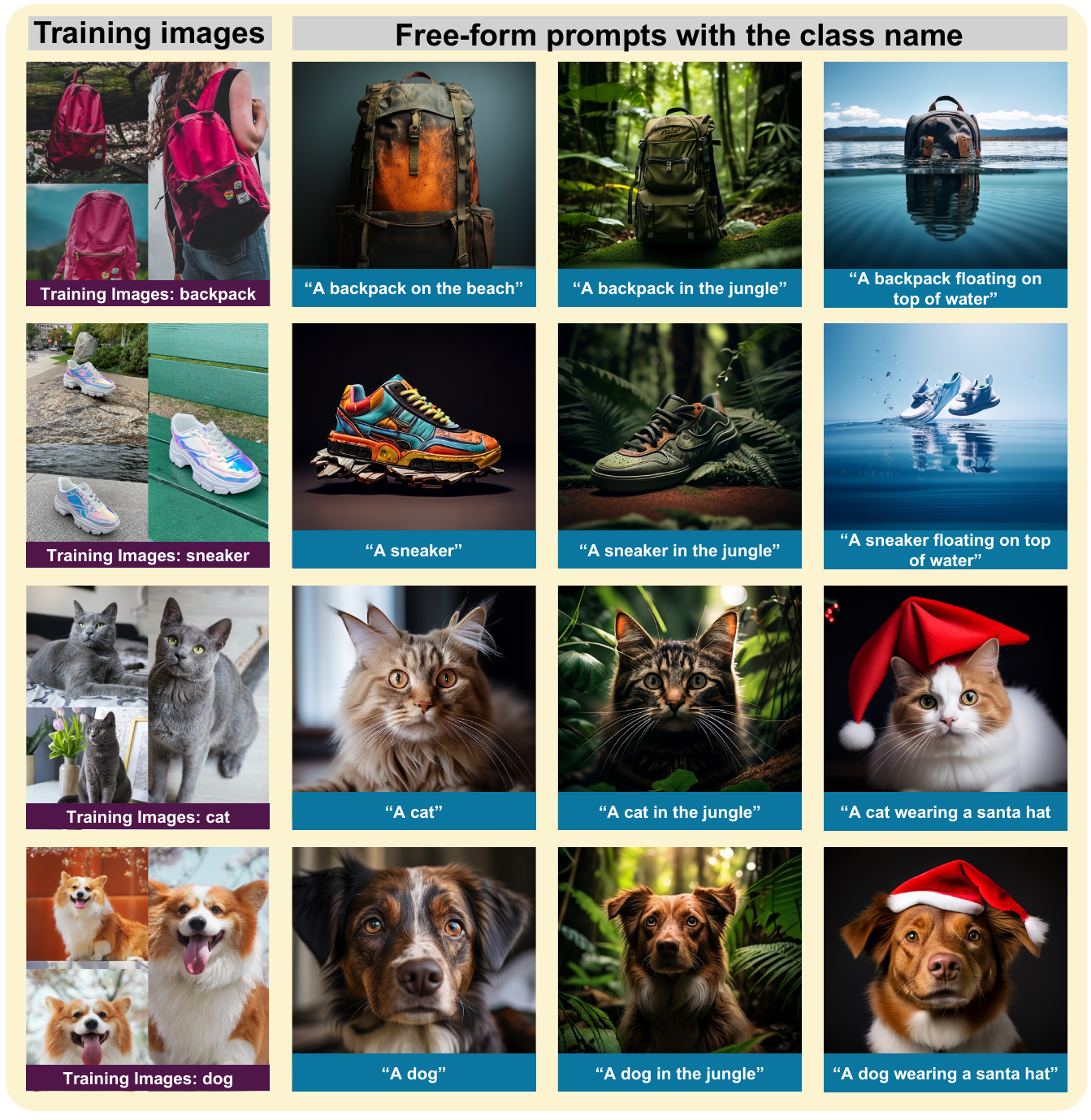}
   \vspace{-10pt}
   \caption{\textbf{Preservation of class semantic priors.} Fine-tuning auto-regressive models with a set of reference images does not result in language drift or reduced output diversity. The first column displays the training images, the next three columns show images generated using free-form prompts that include the specific subject class name.}
   \label{fig:class_preserve}
\end{figure}

\noindent \textbf{Class-Prior ablation.}
As noted in DreamBooth~\cite{ruiz2023dreambooth}, fine-tuning all layers of a diffusion model can lead to issues of \textit{language drift}~\cite{lee2019countering,lu2020countering} and \textit{reduced output diversity}. This occurs because fine-tuning a pre-trained diffusion model on a small set of images causes it gradually forget how to generate subjects of the same class as the target subject, while also reducing the diversity of its outputs.
However, when fine-tuning all layers of the transformer within an auto-regressive model, we do not observe these issues. As illustrated in Figure~\ref{fig:class_preserve}, we present four examples, each in a row. After fine-tuning the model with the images in the first column using the prompt ``A photo of [V] [class\_name]'', the model can still generate diverse and distinct images when prompted with ``A photo of a [class\_name]'' or ``A photo of a [class\_name] in the jungle'', showing no significant resemblance to the training images.

\noindent We investigate the impact of removing the class name in the training prompt by fine-tuning with prompts such as ``A photo of [V]''. The quantitative results, shown in Table~\ref{tab:class_preserve}, indicate that both methods yield comparable results, with fine-tuning using the class name performing slightly better. This suggests that auto-regressive models are more robust in preserving their pre-trained knowledge, eliminating the need for including the Prior Preservation Loss during personalization.

\begin{table}[h]
\centering
\resizebox{\columnwidth}{!}{
  \begin{tabular}{lccc}
    \toprule
    Method & DINO $\uparrow$ & CLIP-I $\uparrow$ & CLIP-T $\uparrow$  \\
    \midrule
    Ours (Lumina-mGPT) w/o class name  & 0.668 & \textbf{0.785} & 0.312\\
    Ours (Lumina-mGPT) & \textbf{0.671} & \textbf{0.785} & \textbf{0.314}\\
    \bottomrule
  \end{tabular}
}
\caption{\textbf{Comparison of subject fidelity and prompt following scores.} Training with and without subject class names in prompts.}
\label{tab:class_preserve}
\end{table}

\begin{table}[tbp]
\centering
\resizebox{\linewidth}{!}{%
\begin{tabular}
{l | l | r | ccc }
\toprule 
\multirow{2}{*}{Rank} & Every N  & \# Trainable & \multirow{2}{*}{DINO $\uparrow$} & \multirow{2}{*}{CLIP-I $\uparrow$} & \multirow{2}{*}{CLIP-T $\uparrow$}\\
& Layer& Parameters & & &  \\

\cmidrule(lr){1-3} \cmidrule(lr){4-6} 

\multirow{3}{*}{$r=16$} & $N=1$ & 12.6M & 0.657 & 0.781 & 0.316 \\
 & $N=2$ & 6.3M & 0.640 & 0.773 & 0.316 \\
 & $N=4$ & 3.2M & 0.630 & 0.769 & \textbf{0.319} \\
 
 \cmidrule(lr){1-6}
 
\multirow{3}{*}{$r=64$} & $N=1$ & 50.4M & 0.657 & 0.778 & 0.312 \\
 & $N=2$ & 25.2M & 0.656 & 0.777 & 0.315 \\
 & $N=4$ & 12.6M & 0.638 & 0.769 & 0.317 \\

 \cmidrule(lr){1-6}
 
\multirow{3}{*}{$r=256$} & $N=1$ & 201.4M & 0.668 & 0.785 & 0.312 \\
 & $N=2$ & 100.7M & 0.654 & 0.775 & 0.315 \\
 & $N=4$ & 50.4M & 0.640 & 0.772 & 0.317 \\

 \cmidrule(lr){1-6}

 \multicolumn{2}{l|}{Full fine-tune} & 1610.6M & \textbf{0.671} & \textbf{0.785} & 0.314 \\ 

\bottomrule
\end{tabular}%
}
\caption{\textbf{Quantitative results comparison on Dreambench~\cite{ruiz2023dreambooth} under different training configurations of Lumina-mGPT~\cite{liu2024lumina-mgpt}.} 
We compare subject fidelity (DINO, CLIP-I) and prompt following (CLIP-T) scores 
across different LoRA ranks $r$ and varying number of training layers. $N$ denotes the interval at which trainable layers are applied, with one trainable layer every $N$ layers.
}
\label{tab:lora}
\end{table}

\noindent \textbf{Fine-tuning transformer with LoRA.}
We fine-tune the transformer layers using LoRA~\cite{hu2022lora} with different LoRA ranks, ranging from 16 to 256. We turn on LoRA for one layer every $N$ layers from the total 32 self-attention layers, where $N$ can be set to 1, 2, or 4. 
The LoRA layers are attached to the projection matrices of the query, key, and value features.  
After optimizing the unique identifier for each subject in the first stage, we finetune the transformer layers with LoRA for about 100 $\sim$ 170 steps in the second stage. During inference, we maintain a fixed CFG value of 4.0 and set the image top-k to 2000.  

\noindent As shown in Table~\ref{tab:lora}, subject fidelity improves with higher LoRA ranks ($r$$\uparrow$) and more trainable layers ($N$$\downarrow$). Although there is a trade-off between subject fidelity and prompt following, prompt following remains strong across different training configurations.

\noindent When the number of trainable layers is fixed, increasing the LoRA rank leads to better subject fidelity. Similarly, for a given rank, training more layers enhances subject fidelity. Generally, more trainable parameters correlate with improved subject fidelity, but there are exceptions. For instance, increasing the rank or trainable parameters but training fewer layers can degrade subject fidelity. This suggests that training more layers has a greater impact on performance than purely increasing the LoRA rank or trainable parameters. It is also indicated in Table~\ref{tab:lora} that fine-tuning parameters within all the transformer layers results in better performance than fine-tuning with LoRA. 

\noindent \textbf{Optimizing only text embeddings.}
We assess the model's performance by optimizing only the text embeddings, without fine-tuning the transformer layers. The quantitative results in Table~\ref{tab:model_layer} show that subject fidelity is significantly lower when training only the embeddings than training both the embeddings and the transformer layers. Since the dimension of a single token embedding is limited to 4096, training only the embeddings is insufficient to capture the complex appearance of the subject. Therefore, fine-tuning the full model is necessary for optimal performance.

\begin{figure}[htbp]
   \centering
   \includegraphics[width=\linewidth]{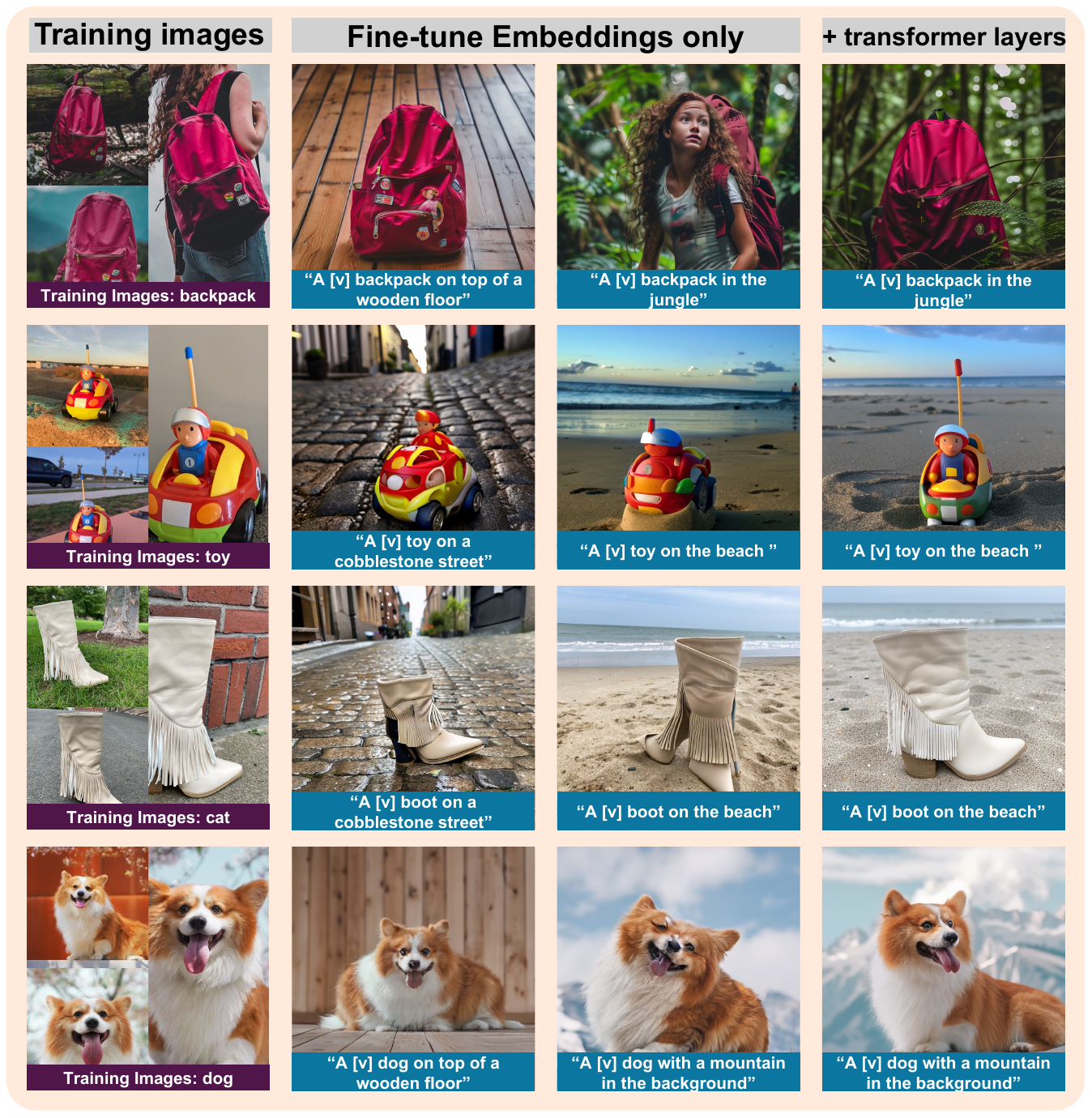}
   \vspace{-10pt}
   \caption{\textbf{Qualitative comparison of fine-tuning strategies: text embeddings only vs. text embeddings and transformer layers.} The first column shows the input images. The second and third columns display images generated by models fine-tuned only on text embeddings, while the fourth column shows results from models fine-tuned on both text embeddings and transformer layers.}
   \label{fig:model_layer}
\end{figure}

\noindent The qualitative comparison is shown in Figure~\ref{fig:model_layer}. The second and third columns display images generated by models fine-tuned only on text embeddings. Although these models are able to capture basic elements of the subject, like color, shape, and texture, they struggle to reproduce those fine details that define the subject's identity, often resulting in unrealistic or distorted patterns. For example, in the third and fourth rows, the boot has two tips, and the dog appears with two bodies. Additionally, due to the limited capacity of text embeddings, the subject sometimes fails to appear as the main focus in the image, as seen in the first row where the backpack is nearly invisible.

\begin{table}[h]
\centering
\resizebox{\columnwidth}{!}{
  \begin{tabular}{lccc}
    \toprule
    Method & DINO $\uparrow$ & CLIP-I $\uparrow$ & CLIP-T $\uparrow$  \\
    \midrule
    Ours (Lumina-mGPT) w/o transformer layers  & 0.601 & 0.754 & \textbf{0.320}\\
    Ours (Lumina-mGPT) w/ transformer layers& \textbf{0.671} & \textbf{0.785} & 0.314\\
    \bottomrule
  \end{tabular}
}
\caption{\textbf{Comparison of subject fidelity and prompt following score.} Training with and without transformer layers.
\label{tab:model_layer}}
\end{table}

\begin{figure}[htbp]
   \centering
   \includegraphics[width=\linewidth]{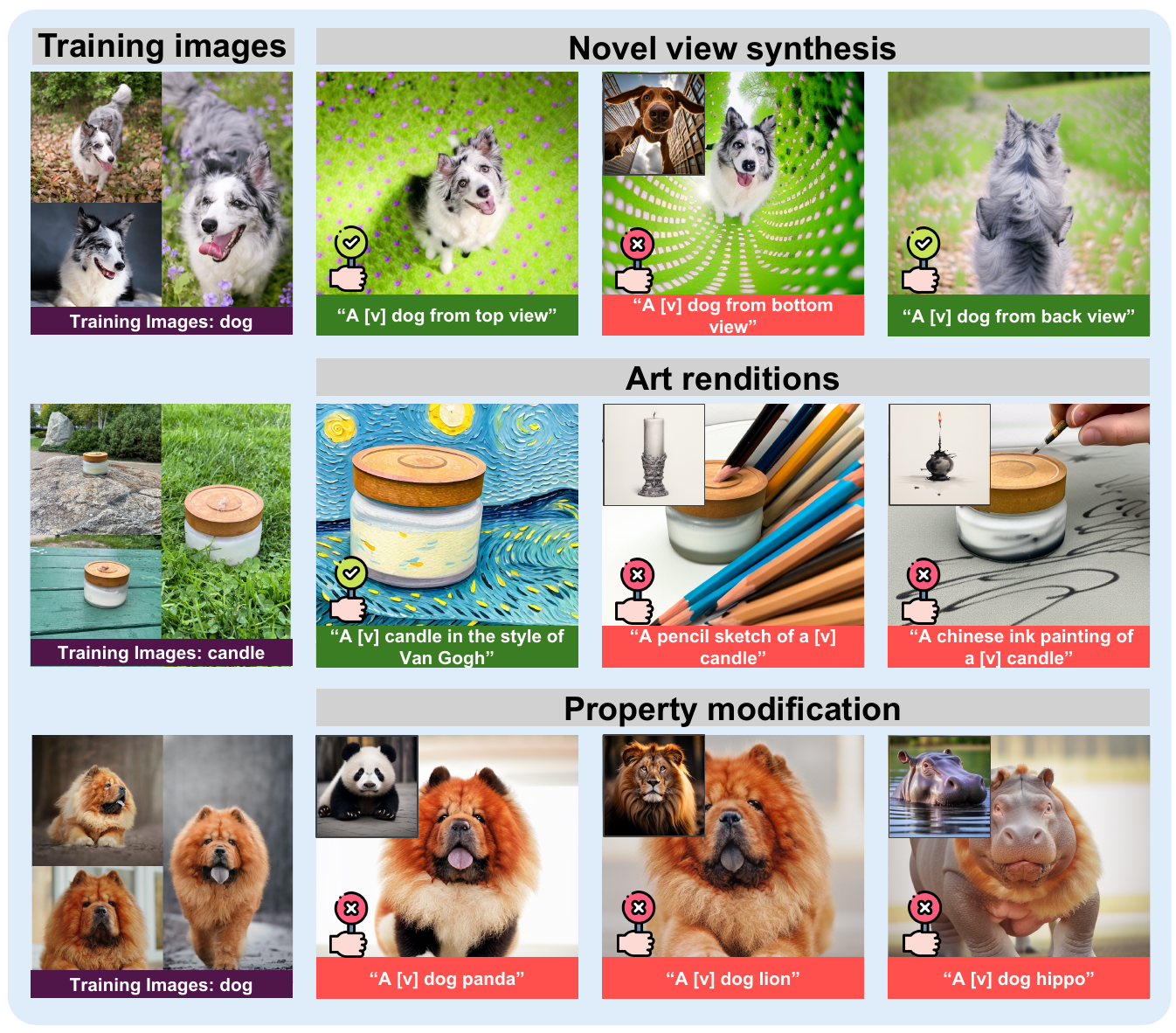}
   \vspace{-10pt}
   \caption{\textbf{Failure cases of various applications.} This figure presents applications of novel view synthesis, artistic renditions, and property modifications. The first column displays the input images. 
   Symbols in the bottom-left corner indicate whether the generated images accurately reflect the prompts. 
   For the failure cases, we include comparison images generated with the same prompt but without the ``[v]'' identifier, allowing us to assess the model's inherent capabilities alongside the effects of fine-tuning.
   }
   \label{fig:limitations}
\end{figure}
\subsection{Limitations}
Our model demonstrates capabilities in re-contextualization, accessorization, and simple property modifications such as color and shape. Beyond these, we aim to explore additional applications. However, results indicate that auto-regressive models struggle with complex scenarios requiring extensive prior knowledge or deep integration of multiple concepts.
Failure cases are illustrated in Figure~\ref{fig:limitations}. A symbol in the bottom-left corner of each image indicates whether the generated image aligns with the prompt. For comparison, we also generate an image using the same prompt, but without the ``[v]'' identifier, which is positioned in the top-left corner.

\noindent \textbf{Novel view synthesis}. The top row of Figure~\ref{fig:limitations} showcases attempts to generate images of the dog from novel viewpoints. The model is tasked with generating perspectives of the specific dog it has never encountered (e.g., top, bottom, or back views). While the model extrapolates class knowledge to somewhat successfully generate top and back views, it fails to produce a correct bottom view due to overfitting to the input images. Although the back view is generated, flaws in the dog's rear make the result appear unnatural.

\noindent \textbf{Art renditions}. The middle row of Figure~\ref{fig:limitations} displays the model's attempts to produce artistic renditions of the object. While the model successfully transfers a candle into Van Gogh's style, it fails with the other two styles. In the pencil sketch and Chinese ink painting examples, the model recognizes the styles but misinterprets them as objects rather than applying them to the subject. This issue may stem from the token-based nature of the auto-regressive model, which associates the identifier ``[v]'' with image tokens of the input subject. Artistic renditions, however, requires replacing the subject with entirely different tokens. As a result, the model circumvents the challenge by reflecting the keywords as objects rather than applying the intended styles.

\noindent \textbf{Property modification}. Figure~\ref{fig:object} illustrates some cases of property modification. The model performs well with simple tasks, such as altering color or shape. However, it struggles with more complex feature combinations. For example, in the top-left corners of the bottom row of Figure~\ref{fig:limitations}, the model fails to merge features of two animals (e.g., ``a dog panda'' or ``a dog lion''). Instead, it generates the second animal mentioned in the prompt only, completely omitting the ``dog''.
When the prompt aims to combine features of the specific Chow Chow dog with another species, the model partially incorporates elements of the dog but fails to create a cohesive fusion. This limitation mirrors the challenges seen in artistic renditions, highlighting the model's lack of flexibility in handling advanced feature integration.

\section{Conclusion}
In this paper, we demonstrate the potential of auto-regressive models for personalized image synthesis through a two-stage training strategy, first optimizing text embedding and then fine-tuning transformer. Our approach achieves comparable subject fidelity and prompt following to the state-of-the-art stable diffusion-based methods such as DreamBooth~\cite{ruiz2023dreambooth}. However, auto-regressive models are slow, taking minutes to generate images, and the fine-tuning also requires 15-20 minutes, limiting real-time applicability. Additionally, the ability to create personalized images raises ethical concerns, such as misuse for misleading content, a challenge common to all generative models. Future work should focus on improving efficiency, addressing ethical risks, and ensuring responsible advancements in personalized generative technologies.
{
    \small
    \bibliographystyle{ieeenat_fullname}
    \bibliography{main}
}

\end{document}